%% file: main.tex
\documentclass[11pt]{article}

\usepackage[final]{acl}

\usepackage{times}
\usepackage{latexsym}
\usepackage[T1]{fontenc}

\usepackage[utf8]{inputenc}

\usepackage{microtype}

\usepackage{inconsolata}

\usepackage{graphicx}
\usepackage{amsmath}
\usepackage{amssymb} 
\usepackage{amsthm}
\usepackage{booktabs} 
\usepackage{multirow}

\title{AdapShot: Adaptive Many-Shot In-Context Learning with Semantic-Aware KV Cache Reuse}

\author{
Jie Ou$^1$, Jinyu Guo$^{1*}$, Shiyao Guo$^1$, Yuang Li$^1$, Ruiqi Wu$^1$, Zhaokun Wang$^1$,\\ 
{\bf Wenyi Li$^1$, Wenhong Tian$^{1*}$} \\
$^1$ School of Information and Software Engineering, \\
        University of Electronic Science and Technology of China\\
        oujieww6@gmail.com, guojinyu@uestc.edu.cn, tian\_wenhong@uestc.edu.cn\\
        Corresponding authors: Jinyu Guo \& Wenhong Tian
}

\begin{document}
\maketitle
\begin{abstract}

Many-Shot In-Context Learning (ICL) has emerged as a promising paradigm, leveraging extensive examples to unlock the reasoning potential of Large Language Models (LLMs). However, existing methods typically rely on a predetermined, fixed number of shots. This static approach often fails to adapt to the varying difficulty of different queries, leading to either insufficient context or interference from noise. Furthermore, the prohibitive computational and memory costs of long contexts severely limit Many-Shot's feasibility. To address the above limitations, we propose \textbf{AdapShot}, which dynamically optimizes shot counts and leverages KV cache reuse for efficient inference. Specifically, we design a probe-based evaluation mechanism that utilizes output entropy to determine the optimal number of shots. To bypass the redundant prefilling computation during both the probing and inference phases, we incorporate a semantics-aware KV cache reuse strategy. Within this reuse strategy, to address positional encoding incompatibilities, we introduce a decoupling and re-encoding method that enables the flexible reordering of cached key-value pairs. Extensive experiments demonstrate that AdapShot achieves an average performance gain of $\sim$10\% and a 4.64$\times$ speedup compared to state-of-the-art DBSA.
\end{abstract}

\input{sections/intro}
\input{sections/related}

\input{sections/pre}

\input{sections/methods}

\input{sections/exp}
\input{sections/conc}

\bibliography{custom}

\end{document}

%% file: sections/intro.tex
\section{Introduction}
In recent years, Large Language Models (LLMs) have achieved milestone breakthroughs in natural language processing. From GPT-3~\cite{brown2020language} to the LLaMA~\cite{touvron2023llama} series, the reasoning capabilities of large language models are continuously being explored and enhanced~\cite{zhang2026tda, zhang2026learning}. And LLMs have demonstrated remarkable In-Context Learning (ICL) capabilities, enabling models to adapt to downstream tasks without parameter updates.

\begin{figure}[t]
\centering
\includegraphics[width=\linewidth]{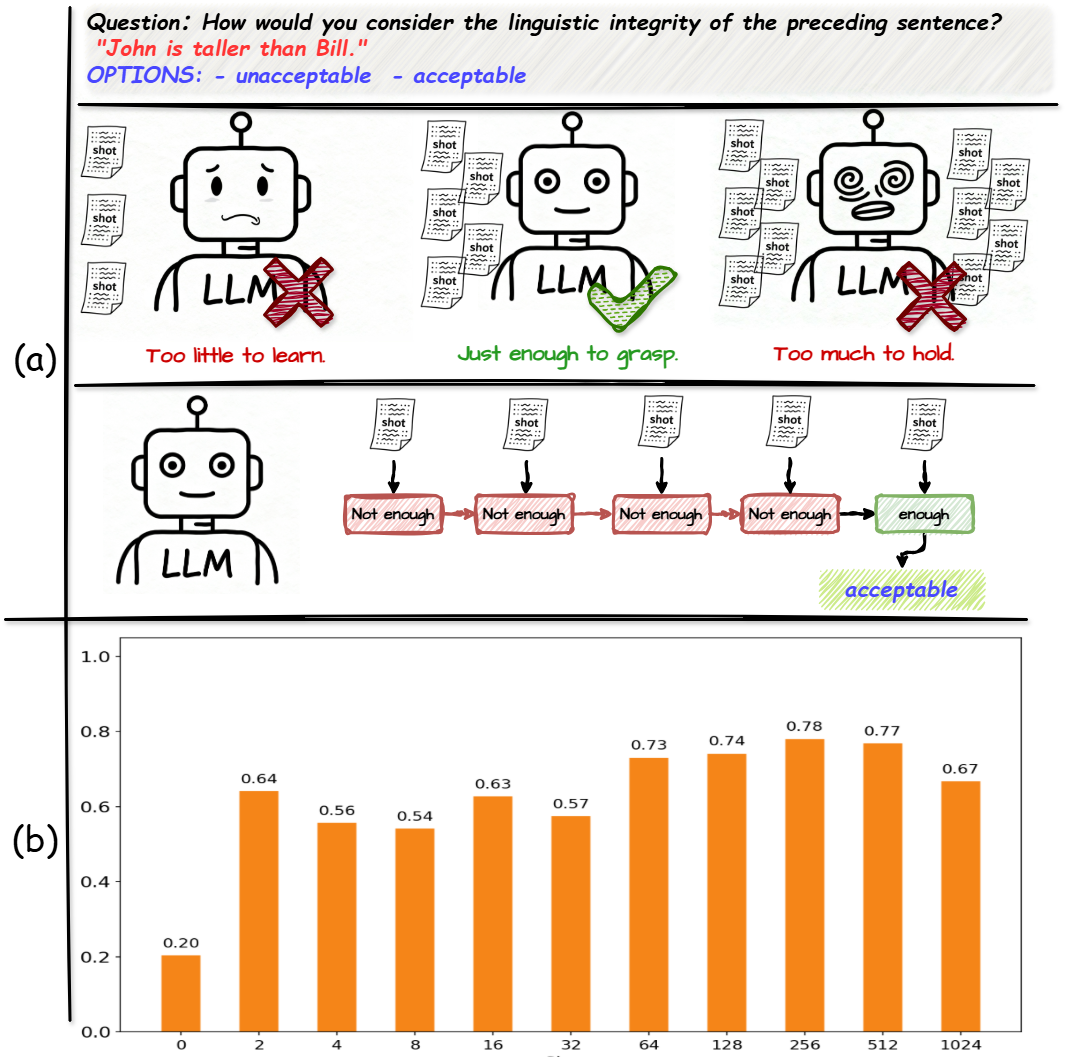}
\caption{(a) Comparison of Few-Shot, adaptive Many-Shot, and Many-Shot approaches. (b) The EM accuracy on the QNLI dataset under different shot settings.}
\label{fig:1}
\end{figure}

While traditional Few-Shot ICL typically relies on fewer than 10 examples, the latest research paradigm is gradually evolving toward \textbf{Many-Shot ICL}~\cite{agarwal2024many,li2023context,bertsch2025context}. Studies show that extending the context window to include hundreds or even thousands of examples can significantly unlock the model's potential in complex reasoning and knowledge-intensive tasks. As illustrated in Figure\ref{fig:1}(b), model accuracy overall improves substantially as the number of examples increases. This scaled ICL paradigm powerfully demonstrates the strong capability of LLMs to perform pattern recognition and reasoning through large-scale contexts.

Many-shot ICL has demonstrated remarkable performance, even rivaling fine-tuning methods in certain scenarios~\cite{yin2024deeper,agarwal2024many}. Recent studies~\cite{bhope2025optiseq,hehierarchical} have further enhanced its effectiveness by exploring and optimizing the ordering strategies of shots within prompts. However, existing research typically employs a predetermined fixed number of shots rather than dynamically adjusting based on actual task requirements. As illustrated in Figure \ref{fig:1}(a), this fixed strategy exhibits significant limitations. When the preset number of shots is insufficient (degrading to a few-shot regime), it leads to inadequate learning of complex tasks by Large Language Models (LLMs). Conversely, when the preset number is excessive, it may exceed the LLM's comprehension capacity or introduce irrelevant noise, thereby interfering with accurate task understanding. 
As shown in Figure \ref{fig:1}(b), more shots do not always yield better results, indicating that different queries require varying numbers of shots rather than following a more-is-better principle. 

On the other hand, regardless of the number of examples used, many-shot ICL faces severe efficiency challenges when fully adopting the large-scale in-context learning paradigm. Furthermore, due to the O($n^2$) time complexity of the self-attention mechanism in the Transformer architecture~\cite{vaswani2017attention}, inference latency increases dramatically as context length grows. This poses significant challenges to GPU memory capacity in practical deployment. Consequently, the efficiency bottleneck of many-shot ICL has become a critical issue that must be addressed.

To address the above challenges, we propose \textbf{AdapShot}, an \textbf{ada}ptive many-\textbf{shot} ICL approach, which dynamically allocates context budget based on the difficulty of each input query. In further, to alleviate the increasingly severe efficiency issues in many-shot ICL, we introduce KV cache reuse into this domain and propose a semantics-aware KV cache construction and position re-encoding method to break the barrier of context sharing across samples. Specifically, we first design a probe-based dynamic evaluation mechanism. This mechanism inserts probes between context windows and calculates output entropy to quantify model confidence, thereby determining the minimum required context budget.
To deal with the efficiency challenges posed by long contexts in many-shot ICL, we leverage an offline global KV cache that eliminates prefilling overhead during probe evaluation and enables effective KV cache reuse.
Within this module, to flexibly reorder samples without compromising the correctness of position information, we design a position decoupling and re-encoding mechanism. It leverages the mathematical properties of Rotary Position Embedding (RoPE) to dynamically map retrieved KV pairs to new logical positions through low-cost vector rotation operations.
This design allows us to freely reorder and concatenate KV blocks according to semantic importance without expensive model forward passes.

Extensive experiments validate AdapShot's superiority, showing it outperforms DBSA with an average $\sim$10\% performance gain and a 4.64$\times$ speedup. These results confirm that AdapShot effectively optimizes the trade-off between accuracy and efficiency in many-shot ICL.
The main contributions of this paper are summarized as follows:
\begin{itemize}
\item We propose a probe-based dynamic budget allocation mechanism that evaluates task difficulty to optimize shot usage and reduce redundant computation. To the best of our knowledge, this is the first work to explore adaptive many-shot ICL based on the probe.

\item We introduce KV cache reuse into many-shot ICL scenarios to address the increasingly severe efficiency challenges. Furthermore, we propose a position decoupling and re-encoding strategy that leverages the rotational properties of RoPE to enable low-cost, lossless, and flexible reordering of shots.

\item Through comprehensive evaluation on various mainstream LLMs and multiple benchmark datasets, experimental results show that our method significantly reduces inference latency while maintaining LLM reasoning quality.
\end{itemize}

%% file: sections/related.tex
\section{Related Work}
\subsection{Many-Shot In-Context Learning}
\begin{figure*}[h]
    \centering
    \includegraphics[width=\linewidth]{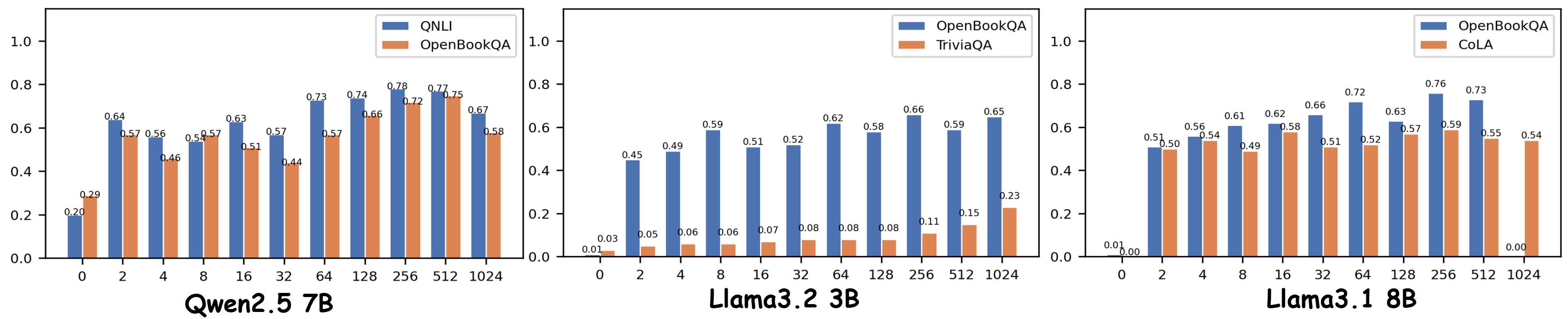}
    \caption{Many-Shot ICL performance of different models across multiple datasets. 
    }
    \label{fig:prelim_analysis}
\end{figure*}

As large language models scale to support extended context windows, in-context learning has evolved from few-shot to many-shot paradigms. \citet{agarwal2024many} pioneered systematic investigation of many-shot ICL, demonstrating substantial performance improvements when incorporating hundreds of demonstration examples. They further explored ``Reinforced ICL'' and ``Unsupervised ICL'' to mitigate dependency on annotated data. \citet{bertsch2025context} revealed continued performance gains with increasing examples, particularly for tasks with large label spaces, while also exhibiting reduced sensitivity to input permutations.

Existing research has explored many-shot ICL optimization across several dimensions. For a demonstration organization, \citet{hehierarchical} proposed HIDO to address order instability. For selection strategies, \citet{golchintowards} reduced computational overhead through similarity-based retrieval with cached demonstrations, while \citet{wanfew} introduced BRIDGE to optimize influential example selection. For scaling approaches, \citet{gu2025scaling} and \citet{chenmaple} developed IterPSD and MAPLE using pseudo-labeling techniques. \citet{zou2025many} categorized ICL into Similar Sample Learning (SSL) and All Sample Learning (ASL), revealing that models handle 64k tokens in SSL but some degrade at 16k tokens in ASL. Applications have extended to specialized domains such as molecular inverse design \cite{moayedpour2024many}.

Despite these advances, current methods still rely on empirically fixed demonstration quantities, face order-sensitivity constraints that disrupt KV cache sharing, and encounter severe efficiency bottlenecks. While \citet{xiao-etal-2025-efficient} enhanced efficiency via dynamic block-sparse attention, validation remains confined to simple NLU tasks. Our AdapShot addresses these challenges by dynamically adjusting the demonstration scale based on query difficulty.

\subsection{Efficient LLM Inference Technology}
Sparse attention methods like block-sparse mechanisms \citet{child2019generating, zaheer2020big, wangprecision, acharyastar} and hierarchical structures \citet{yang2016hierarchical} reduce computational complexity but typically require retraining. \citet{ratner2023parallel} proposed Parallel Context Windows that avoids retraining through block-sparse attention, though it remains limited to context-constrained models.

KV cache compression techniques include token eviction strategies, where \citet{xiaoefficient} discovered "attention sink" phenomena leading to StreamingLLM that preserves only initial and recent KV pairs. Subsequent works \citet{li2024snapkv, zhang2023h2o, liu2025chunkkv} improved selective retention approaches. However, eviction strategies face challenges in many-shot ICL scenarios where different queries require attention to different example subsets. Alternative compression methods like quantization and low-rank approximation \citet{liu2024minicache, zhang2024unifying} reduce memory usage while preserving all tokens. The efficiency of LLMs has also been extensively studied in multimedia and computer vision (\citet{zheng2025joint,cao2026language,zheng2026llava}).

Our work integrates inference acceleration techniques into the many-shot ICL domain, enabling practical deployment of demonstration-rich learning that was previously computationally prohibitive.

%% file: sections/pre.tex
\section{Preliminary Study and Motivation}
In Many-Shot ICL, we maintain a repository $\mathcal{S}=\{(x_1, y_1), \ldots, (x_N, y_N)\}$ of $N$ input-output pairs. For a query $q$, we select $k$ examples (typically hundreds to thousands) from $\mathcal{S}$ to form context $C$. The LLM then processes $\text{LLM}(I \oplus C \oplus q)$ to generate the output, where $I$ is the instruction and $\oplus$ denotes concatenation.

\textbf{Observation 1: The relationship between the number of examples and performance is non-monotonic.} As shown in Figure\ref{fig:prelim_analysis}, as the number of examples increases from 0 to 1024, model performance exhibits complex variation patterns. For instance, Llama-3.1-8B on OpenBookQA shows accuracy dropping from 76\% with 256 examples to 73\% with 512 examples, and crashes due to memory limitations at 1024 examples. Additionally, Qwen2.5-7B on QNLI reaches peak performance of 78\% at 256 examples, followed by a gradual decline. This performance degradation phenomenon indicates that excessive examples may severely interfere with the model's reasoning process.

\textbf{Observation 2: Different models exhibit significant variations in perceiving task complexity.} Notably, "task difficulty" is not an objective universal standard but is highly dependent on the capability boundaries of specific models. For example, Llama-3.2-3B achieves only 23\% accuracy on TriviaQA even with 1024 examples, indicating this task is extremely challenging for 3B-scale models. However, on another knowledge-intensive task, OpenBookQA, the same model achieves approximately 65\% performance with 64-256 examples. Furthermore, the "optimal number of examples" varies dramatically across models: Qwen2.5-7B requires 256-512 examples to reach optimal performance on most tasks, while Llama-3.1-8B's performance on CoLA is relatively insensitive to example count, consistently fluctuating between 50-60\%.

Based on these observations, we argue that the current "one-size-fits-all" static example configuration strategy is fundamentally flawed. Each "model-task-query" triplet requires a unique optimal example configuration. This adaptive approach avoids both computational waste and performance degradation from example overload.

%% file: sections/methods.tex
\begin{figure}[t]
\centering
\includegraphics[width=\linewidth]{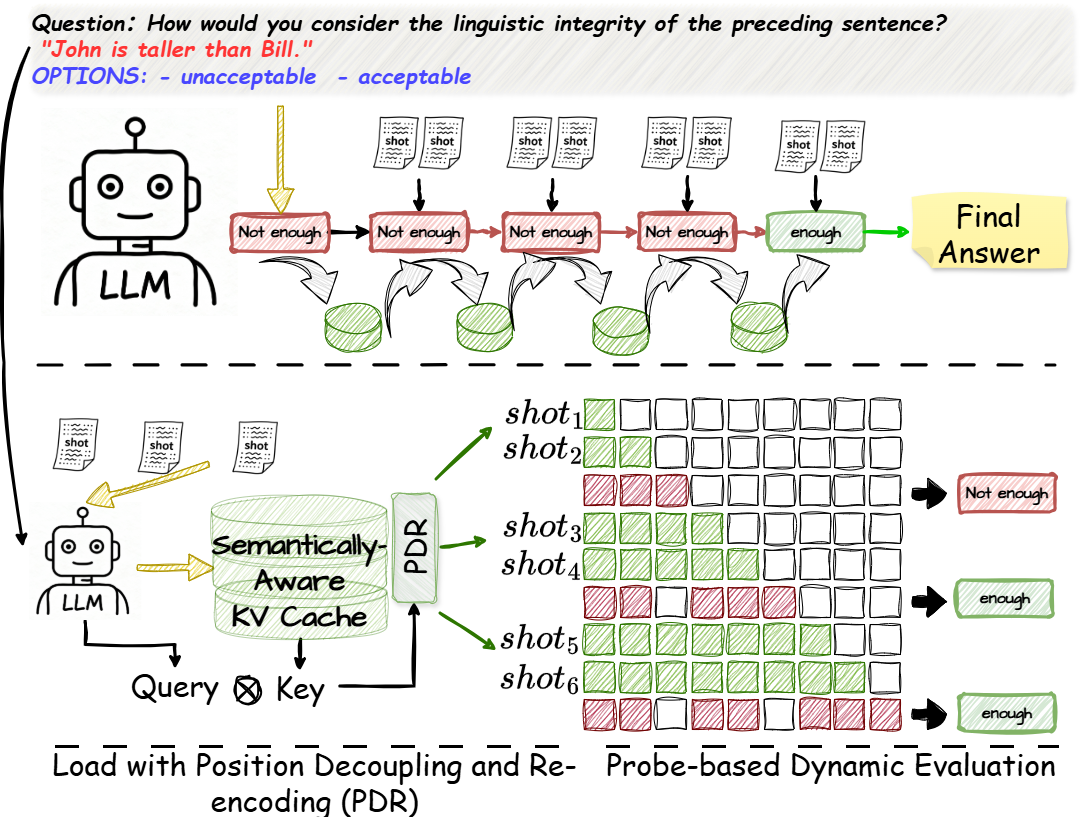}
\caption{The pipeline of AdapShot.}
\label{fig:pipeline}
\end{figure}

\section{Method}
AdapShot provides an adaptive inference framework that dynamically adjusts context scale based on the difficulty of input queries, enabling flexible and efficient allocation of computational resources. As illustrated in Figure~\ref{fig:pipeline}, AdapShot first constructs a semantically-aware global KV cache pool during the offline phase. Subsequently, during the inference phase, it employs a probe-based dynamic evaluation mechanism that estimates model confidence using minimal single-token generation. This allows adaptive determination of the minimum effective example set. For activated examples, a position re-encoding mechanism is utilized to perform concatenation and position correction solely at the KV level, avoiding redundant prefilling computation.

\subsection{Probe-based Dynamic Evaluation}
\label{sec:probe-based}
Many-Shot ICL deployment often falls into the misconception that "more examples are always better," leading to significant memory burden and inference latency, and even performance degradation. To address this, we propose a \textbf{probe-based dynamic evaluation mechanism} to assess the model's confidence on the current query and dynamically determine whether to continue expanding the number of examples.

Before formal probe evaluation, the system first performs semantic relevance ranking on the global example pool 
\(\mathcal{S} = \{\text{shot}_1, \ldots, \text{shot}_N\}\)
based on query \(q\). Following the sorting method in Section \ref{m:sort}, the obtained 
\(\mathcal{S}_{\text{sorted}} = [\text{shot}^{(1)}, \ldots, \text{shot}^{(N)}]\)
ensures that the most relevant examples are prioritized for subsequent procedures.

Based on the sorted results, the probe mechanism adopts an iterative process to dynamically determine the final context scale. Assuming a step size of \(n\) examples per iteration, the \(k\)-th iteration (\(k \ge 1\)) activates the top \((k \times n)\) most relevant examples and constructs a probe context:
\begin{equation}\label{eq:probe_context}
    \text{C}_{k}^{\text{probe}}
    = I \oplus \mathcal{S}_{\text{sorted}}[:k \times n]
    \oplus q \oplus \text{P}_{\text{probe}},
\end{equation}
where \(I\) represents an optional instruction prefix or task description, \(\oplus\) denotes string concatenation, and \(\text{P}_{\text{probe}}\) is a probe prompt (e.g., "Based on the above information, are you confident enough to answer?"). The model then generates only a single token (e.g., "Yes" or "No") conditioned on \(\text{C}_{k}^{\text{probe}}\), and computes the entropy of its output distribution \(\{p(\text{Yes}), p(\text{No})\}\):
\begin{equation}\label{eq:entropy}
    H_k = -\sum_{c \in \{\text{Yes},\, \text{No}\}}
    p\bigl(c \mid \text{C}_{k}^{\text{probe}}\bigr)\,\log p\bigl(c \mid \text{C}_{k}^{\text{probe}}\bigr).
\end{equation}
If \(H_k \le \tau\) (where \(\tau\) is a preset threshold), the model is considered to have sufficient confidence for the task. The iteration stops, and the currently activated example set is used for subsequent formal inference. If \(H_k > \tau\), we set \(k \leftarrow k + 1\) and proceed to the next probe iteration.

To mitigate the increased time overhead from multiple probe cycles, we leverage tree-structured attention for single-round parallel multi-probe verification. Given the accelerator's inherent parallelism, this approach does not introduce significant computational overhead but rather reduces the number of iteration rounds.

For candidate example counts $\{n_1, n_2, \ldots, n_m\}$, we construct the following parallel input sequence:
$[\text{shot}_1, \ldots, \text{shot}_{n_1}, \text{probe}_1, 
\text{shot}_{n_1+1}, \ldots, \text{shot}_{n_2}, \\ \text{probe}_2, \ldots]$
Additionally, we construct a tree-structured attention mask $M$ such that each probe $\text{probe}_i$ only attends to its corresponding first $n_i$ examples:
\begin{equation}
M_{ij} = \begin{cases}
1 & \text{if } j \le n_k \text{ and } i = \text{probe}_k \\
0 & \text{otherwise}
\end{cases}
\end{equation}

This allows each probe $\text{probe}_i$ to independently generate confidence assessments based on different amounts of context in a single forward pass, while the overall computational complexity is comparable to processing the longest sequence. Subsequently, the minimum number of examples satisfying the confidence threshold is selected as the optimal configuration.

\subsection{Semantically-Aware KV Cache}
\label{m:sort}
Many-Shot ICL requires prefilling hundreds or even thousands of examples during inference. The Key-Value (KV) representations of these examples introduce memory redundancy and additional inference latency. To substantially reduce this computational overhead without compromising model performance, AdapShot proposes a semantically-aware hierarchical KV cache for maximum cross-sample reuse.

During the offline precomputation phase, AdapShot performs one-time offline computation on the global example pool \(\mathcal{S}\) to obtain the KV representations for each example \(\text{shot}_i\) across all layers \(\ell \in [1, L]\) and attention heads \(h \in [1, H]\):
\begin{equation}\label{eq:prefill}
\begin{split}
        \Bigl(\mathcal{K}_{\ell,h}^{(i)},\, \mathcal{V}_{\ell,h}^{(i)}\Bigr)
    = \text{Prefill}\bigl(\text{shot}_i\bigr),\\
    \quad
    \forall\, i \in [1,N],\ \ell \in [1,L],\ h \in [1,H],
\end{split}
\end{equation}
where \(L\) denotes the number of model layers, \(H\) represents the number of attention heads per layer, and \(\mathcal{K}_{\ell,h}^{(i)}\) and \(\mathcal{V}_{\ell,h}^{(i)}\) are the Key and Value tensors for example \(\text{shot}_i\) at layer \(\ell\) and head \(h\), respectively. These KV vectors are organized and stored in a hierarchical structure within the global cache pool.

During online inference, for a new query \(q\), AdapShot first encodes \(q\) to obtain Query vectors across all layers and heads:
\(\mathbf{Q}_{\ell,h}^{(q)} \in \mathbb{R}^{T_q \times d}\):
\begin{equation}\label{eq:query_encode}
    \mathbf{Q}_{\ell,h}^{(q)} = \text{Encode}_{\ell,h}(q),
\end{equation}
where \(T_q\) denotes the number of tokens in the query sequence and \(d\) is the hidden dimension.

Next, the system computes semantic relevance between the query vectors and the Key vectors of all examples in the global pool. Specifically, for example \(\text{shot}_i\), we calculate the attention scores between the query and the example at layer \(\ell\) (select a representative layer for different LLMs, e.g. the last layer) and head \(h\):
\begin{equation}\label{eq:attention_score}
    \mathbf{A}_{\ell,h}^{(i)} = \text{softmax}\left(\frac{\mathbf{Q}_{\ell,h}^{(q)} \cdot (\mathcal{K}_{\ell,h}^{(i)})^T}{\sqrt{d}}\right),
\end{equation}
where \(\mathbf{A}_{\ell,h}^{(i)} \in \mathbb{R}^{T_q \times T_i}\) represents the token-level attention matrix between the query sequence and example \(i\), with \(T_i\) being the number of tokens in example \(i\). We average the attention scores across all tokens to obtain the relevance score $s^{(i)}$ for example $i$. We rank all examples by their relevance scores $\{s^{(1)}, s^{(2)}, \ldots, s^{(N)}\}$ in descending order and selects the top $k$ examples as the active set $\mathcal{S}_{\text{active}}$. This attention-based retrieval identifies the most relevant examples for each query without additional encoding models, while reusing the corresponding KV cache.

\subsection{Position Decoupling and Re-encoding}
\label{sec:rope-compensation}

During offline construction, each example $\text{shot}_i$ is prefilled independently with position encodings starting from 0. However, when examples are reordered during online inference based on relevance scores, directly concatenating cached KV pairs causes position conflicts.

Consider two examples $\text{shot}_A$ and $\text{shot}_B$ with lengths $T_A$ and $T_B$. During offline prefilling, their position indices are $[0, 1, \ldots, T_A-1]$ and $[0, 1, \ldots, T_B-1]$ respectively. When concatenating $\text{shot}_B$ after $\text{shot}_A$, the expected positions should be $[0, \ldots, T_A-1, T_A, \ldots, T_A+T_B-1]$, but $\text{shot}_B$'s cached Keys still encode positions starting from 0, resulting in a position offset $\Delta = T_A$.

\begin{table*}[h]
\centering
\caption{Performance comparison (Exact Match) of AdapShot and baselines across LLaMA and Qwen architectures. Best results are bolded. O.O.M. denotes Out of Memory. We extended DBSA$^\dag$ to these datasets.}
\label{tab:main_accuracy}
\resizebox{\textwidth}{!}{
\begin{tabular}{l|ccc|ccccccc}
\toprule
\multirow{2}{*}{\textbf{Method}} & \multicolumn{3}{c|}{\textbf{LLaMA-3.2 (3B)}} & \multicolumn{7}{c}{\textbf{Qwen2.5-7B}} \\
\cmidrule{2-11}
 & CoLA & QNLI & PIQA & CoLA & QNLI & PIQA & SQuAD v2 & SVAMP & GSM8K & MathQA \\
\midrule
Zero-shot & 0.421 & 0.508 & 0.107 & 0.169 & 0.203 & 0.298 & 0.095 & 0.136 & 0.456 & 0.012 \\
Few-shot (8) & 0.449 & 0.517 & 0.193 & 0.484 & 0.541 & 0.318 & 0.089 & 0.107 & 0.305 & 0.350 \\
Many-shot (256) & 0.528 & 0.557 & 0.345 & 0.488 & 0.780 & 0.560 & 0.309 & 0.146 & 0.450 & 0.483 \\
Many-shot (512) & 0.540 & 0.579 & \textbf{0.473} & 0.542 & 0.768 & 0.522 & 0.236 & 0.456 & 0.581 & 0.232 \\
Many-shot (1024) & 0.533 & 0.526 & O.O.M. & 0.550 & 0.667 & O.O.M. & O.O.M. & O.O.M. & O.O.M. & O.O.M. \\
DBSA$^\dag$ & 0.649 & \textbf{0.619} & 0.368 & 0.657 & 0.807 & 0.404 & 0.404 & 0.737 & 0.343 & 0.323 \\
\midrule
\textbf{AdapShot (Ours)} & \textbf{0.777} & 0.512 & 0.469 & \textbf{0.699} & \textbf{0.825} & \textbf{0.599} & \textbf{0.465} & \textbf{0.790} & \textbf{0.649} & \textbf{0.524} \\
\bottomrule
\end{tabular}
}
\end{table*}
\begin{table*}[h]
\centering
\caption{Inference speedup of AdapShot relative to Many-shot baselines and DBSA on Qwen2.5-7B. Speedup is defined as the ratio of the baseline's average latency to AdapShot's average latency. Higher is better.}
\label{tab:main_efficiency}
\resizebox{0.75\linewidth}{!}{
\begin{tabular}{l|cccc|c}
\toprule
\textbf{Dataset} & \textbf{vs. 64-shot} & \textbf{vs. 128-shot} & \textbf{vs. 256-shot} & \textbf{vs. 512-shot} & \textbf{vs. DBSA} \\
\midrule
CoLA & 2.18$\times$ & 3.03$\times$ & 3.41$\times$ & 3.23$\times$ & 4.62$\times$ \\
QNLI & 3.42$\times$ & 2.85$\times$ & 2.98$\times$ & 4.63$\times$ & 5.63$\times$ \\
PIQA & 2.08$\times$ & 2.15$\times$ & 2.63$\times$ & 3.54$\times$ & 4.27$\times$ \\
SQuAD v2 & 1.72$\times$ & 1.65$\times$ & 1.77$\times$ & 4.10$\times$ & 2.64$\times$ \\
SVAMP & 2.09$\times$ & 2.93$\times$ & 3.49$\times$ & 4.53$\times$ & 4.72$\times$ \\
GSM8K & 3.21$\times$ & 3.41$\times$ & 3.59$\times$ & 7.59$\times$ & 4.27$\times$ \\
MathQA & 2.56$\times$ & 3.58$\times$ & 4.77$\times$ & 9.12$\times$ & 6.33$\times$ \\
\midrule
\textbf{Average} & \textbf{2.47$\times$} & \textbf{2.80$\times$} & \textbf{3.23$\times$} & \textbf{5.25$\times$} & \textbf{4.64$\times$} \\
\bottomrule
\end{tabular}
}
\end{table*}
To address this, we leverage the rotational composability of RoPE. The RoPE encoding formula is:
\begin{equation}\label{eq:rope_encoding}
\begin{split}
    \text{RoPE}(\mathbf{x}, p) =\, & \mathbf{x} \odot \cos(p \boldsymbol{\theta}) \\
    & + \text{rotate\_half}(\mathbf{x}) \odot \sin(p \boldsymbol{\theta}),
\end{split}
\end{equation}
where $\boldsymbol{\theta} = [\theta_0, \theta_1, \ldots, \theta_{d/2-1}]$ with $\theta_i = 10000^{-2i/d}$.
The key property of RoPE is that a vector with position $p_1$ can be transformed to position $p_2$ through an additional rotation of $\Delta = p_2 - p_1$. Thus, for a cached Key vector $\mathcal{K}_{\ell,h}^{(i)}$ at original position $p_{\text{old}}$ that needs to be at position $p_{\text{new}}$, we compute the corrected Key as:
\begin{equation}\label{eq:rope_compensation}
\begin{split}
    \mathcal{K}_{\ell,h, \text{new}}^{(i)} =\, & \mathcal{K}_{\ell,h}^{(i)} \odot \cos(\Delta \boldsymbol{\theta}) \\
    & + \text{rotate\_half}(\mathcal{K}_{\ell,h}^{(i)}) \odot \sin(\Delta \boldsymbol{\theta})
\end{split}
\end{equation}

This mechanism enables efficient position re-encoding without recomputing Keys, completely decoupling the physical storage order from logical positions during inference. This allows AdapShot to reuse cached KV pairs flexibly while maintaining correct attention computation after semantic reordering.

%% file: sections/exp.tex
\section{Experiments}
We employ the CoT Collection dataset~\cite{kim2023cot} for comprehensive evaluation, selecting 7 representative tasks across diverse domains and reasoning types: natural language understanding (CoLA, QNLI, PIQA), question answering (SQuAD v2), and mathematical reasoning (SVAMP, GSM8K, MathQA).

\subsection{Baselines}
We compare AdapShot against multiple baselines for comprehensive evaluation. For performance, we test: Zero-shot (no examples), Few-shot (4-8 examples), Many-shot (64, 128, 256, 512, 1024 examples), and DBSA~\cite{xiao-etal-2025-efficient} (dynamic block sparse attention with cached example groups). For efficiency, we measure speedup (ratio of baseline to our method's inference time) against Many-shot variants and DBSA.

\subsection{Experimental Setup}

All experiments were conducted on a Huawei Ascend 910B2 NPU cluster. The computational resources included 8 Ascend 910B2 NPUs, each equipped with 65,536MB of memory. The CPU was a HUAWEI Kunpeng 920 5250, with 48 cores per socket, and the system was configured with 1.5TB of memory. The software environment was based on the Huawei Cloud EulerOS 2.0 operating system.

\subsection{Performance Analysis}
\label{sec:performance_analysis}

Table~\ref{tab:main_accuracy} presents the performance comparison between AdapShot and baseline methods across LLaMA and Qwen architectures. On the more capable Qwen2.5-7B model, AdapShot demonstrates superior cross-task generalization, outperforming all baselines, including the state-of-the-art DBSA method, across every evaluated dataset. Notably, AdapShot achieves substantial gains in complex reasoning and comprehension tasks, securing 0.790 on SVAMP (vs. 0.737 for DBSA) and 0.465 on SQuAD v2 (vs. 0.404 for DBSA). On LLaMA-3.2 (3B), our method attains an exceptional score of 0.777 on CoLA, representing a 19.7\% improvement over the best DBSA$^\dag$. These results validate that AdapShot effectively surpasses current SOTA approaches by adaptively tailoring retrieval strategies to varying model capabilities.

Furthermore, the results expose the inherent limitations of fixed-length strategies, where increasing context length does not guarantee better performance. As observed in the Qwen2.5-7B results on QNLI, extending the context from 256 to 1024 shots causes performance to degrade from 0.780 to 0.667, indicating that excessive examples introduce detrimental noise. Additionally, the 1024-shot setting frequently triggers Out-Of-Memory failures. In contrast, AdapShot mitigates these issues by dynamically identifying the optimal context budget for each query. It avoids the noise accumulation seen in over-extended contexts while preventing memory overflows, demonstrating a robust balance between computational efficiency and task accuracy that rigid baselines fail to achieve.

\subsection{Efficiency Evaluation}
\label{sec:efficiency_evaluation}

Table~\ref{tab:main_efficiency} presents the inference speedup of AdapShot relative to fixed Many-shot baselines and the DBSA method on Qwen2.5-7B. The results demonstrate that AdapShot achieves substantial improvements in computational efficiency across tasks of varying complexity, with the advantage becoming increasingly pronounced as the context scale expands. In reasoning-heavy benchmarks, AdapShot attains peak speedups of 9.12$\times$ on MathQA and 7.59$\times$ on GSM8K compared to the 512-shot baseline. Even against DBSA, which employs dynamic block-sparse attention, AdapShot maintains a robust average speedup of 4.64$\times$, reaching up to 6.33$\times$ on MathQA. These findings confirm that AdapShot effectively minimizes redundant computation, offering superior scalability in large-context scenarios.

\subsection{Ablation Studies}
\begin{figure}[h]
\centering
\includegraphics[width=\linewidth]{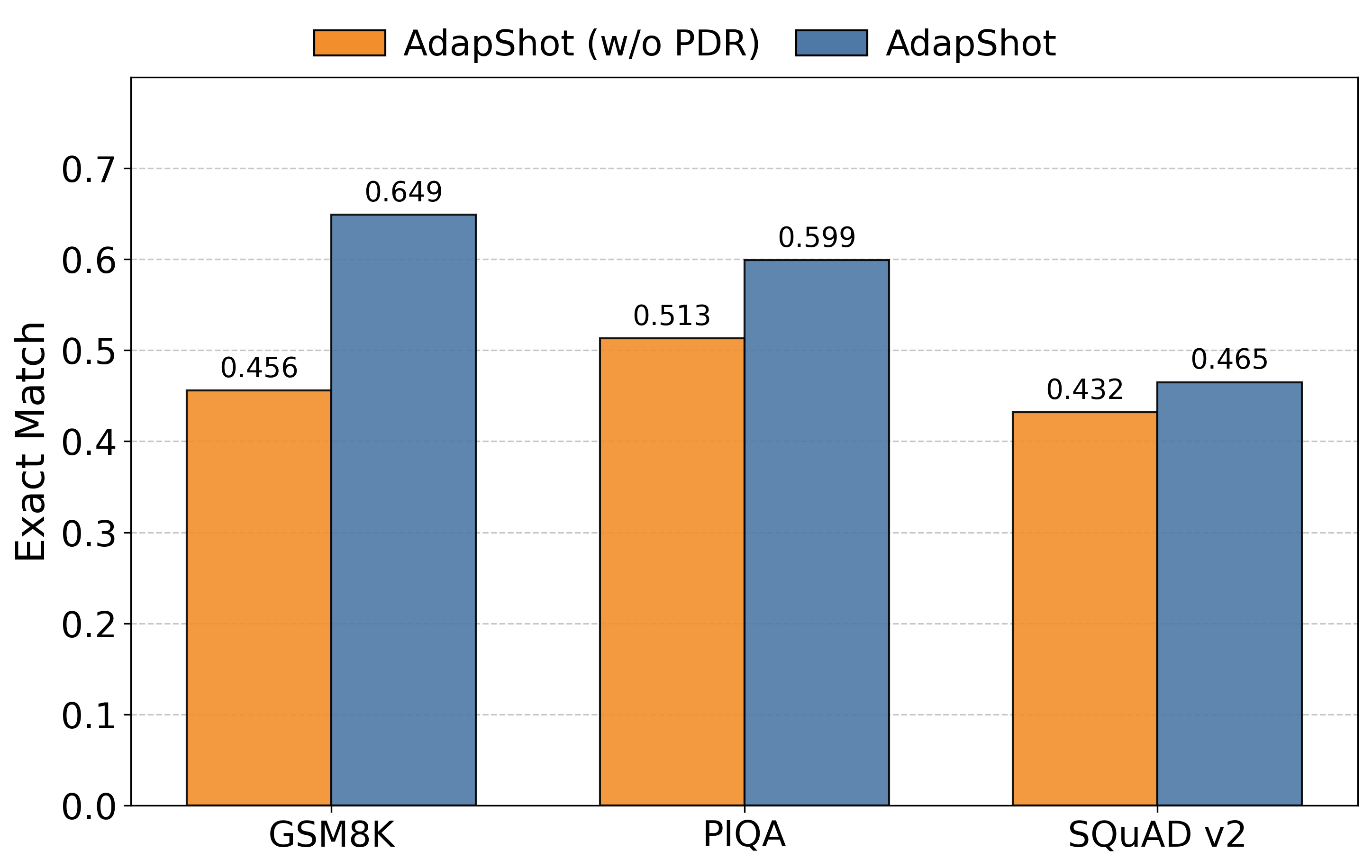}
\caption{Ablation study on Position Decoupling and Re-encoding (PDR) module across different datasets and methods.}
\label{fig:pdr_ablation}
\end{figure}

\noindent\textbf{Effectiveness of PDR:} As illustrated in Figure \ref{fig:pdr_ablation}, removing the PDR module results in substantial performance degradation, with accuracy dropping by 29.7\% on GSM8K (0.649$\to$0.456), 14.4\% on PIQA, and 7.1\% on SQuAD v2. These results underscore the necessity of PDR in correcting position encoding misalignments caused by dynamic example reordering, ensuring the attention mechanism functions correctly during KV cache reuse.

\begin{figure}[h]
\centering
\includegraphics[width=\linewidth]{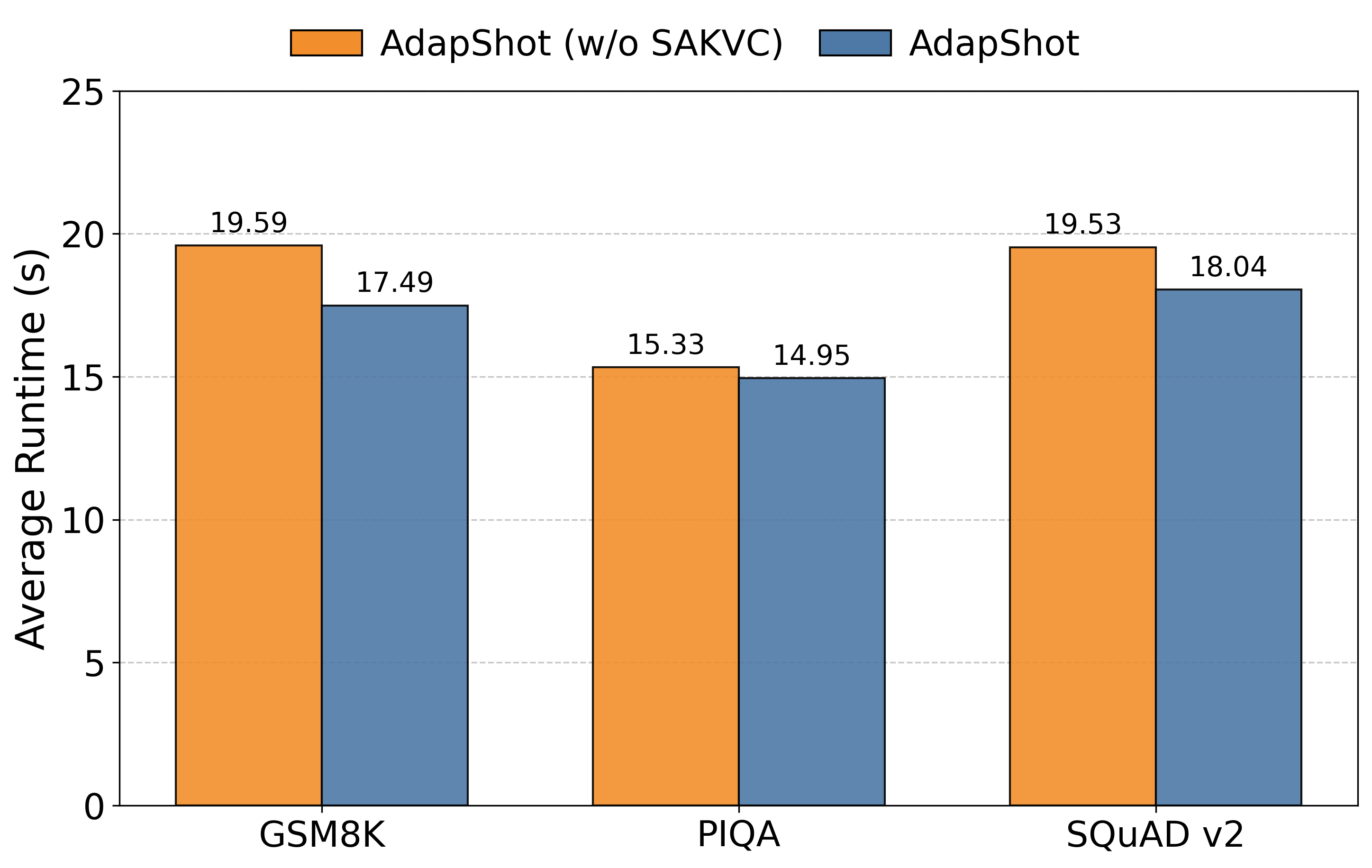}
\caption{Runtime comparison between AdapShot with and without Semantically-Aware KV Cache (SAKVC) across different datasets.}
\label{fig:runtime_comparison}
\end{figure}

\noindent\textbf{Efficiency Gains from SAKVC:} Figure \ref{fig:runtime_comparison} highlights the latency benefits of our semantically-aware kv cache strategy. By bypassing redundant prefilling for activated examples, SAKVC consistently reduces runtime, achieving reductions of 10.7\% on GSM8K (19.59s$\to$17.49s), 7.6\% on SQuAD v2, and 2.5\% on PIQA. This confirms that our SAKVC effectively alleviates the prefilling bottleneck inherent in Many-Shot ICL.

\subsection{Extended Analysis}
\begin{figure}[h]
    \centering
    \includegraphics[width=\linewidth]{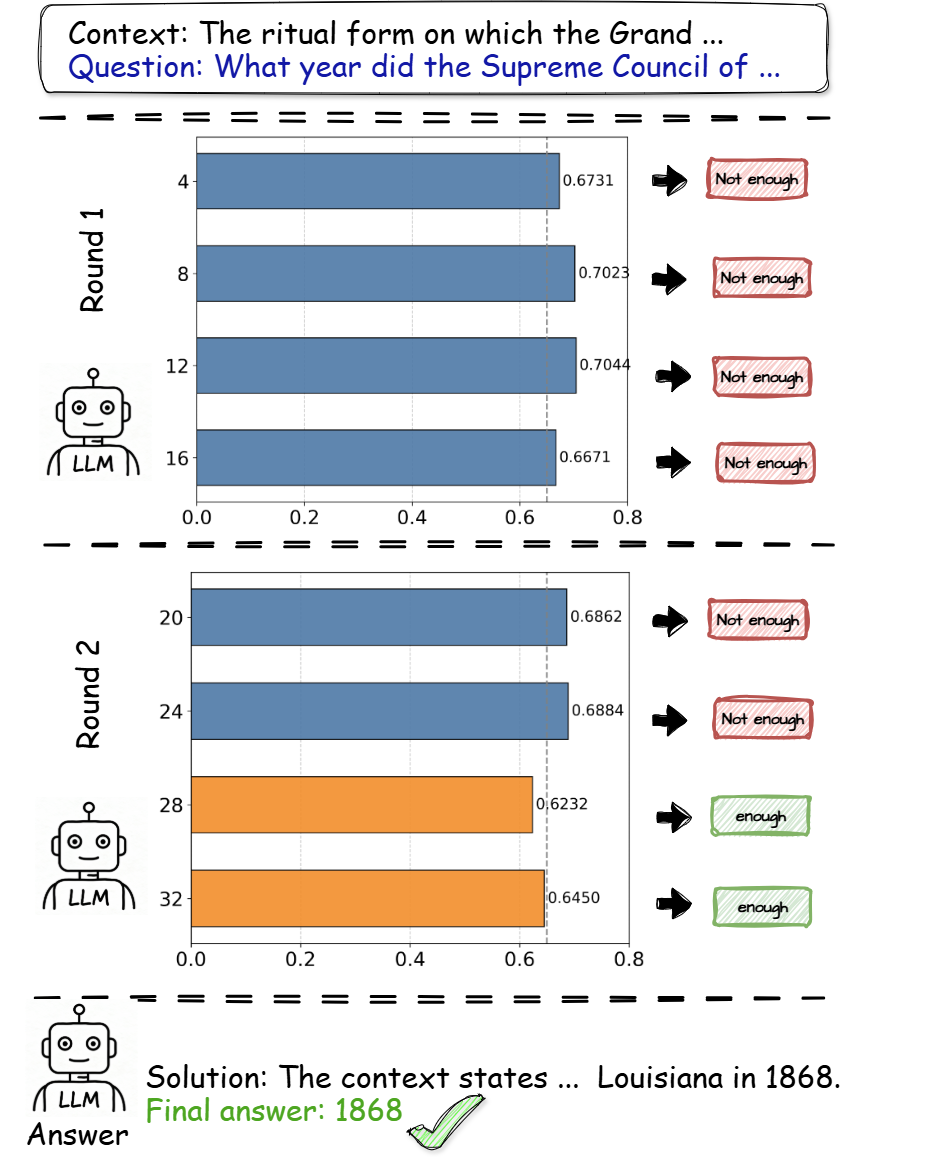}
    \caption{Visualization of AdapShot's dynamic shot selection process on a SQuAD reading comprehension task.}
    \label{fig:case_study}
\end{figure}

\textbf{Case Study:} We visualize AdapShot's dynamic shot selection process and its effectiveness in activating knowledge using Qwen2.5-7B on a SQuAD reading comprehension task that asks:  \textit{What year did the Supreme Council of the Ancient and Accepted Scottish Rite of Louisiana appear in the jurisdiction of the Grand Lodge of Louisiana?} As shown in Figure \ref{fig:case_study}, initially, four parallel probes with 4, 8, 12, and 16 shots produced entropy values of 0.6731, 0.7023, 0.7044, and 0.6671, respectively, all exceeding the 0.65 threshold and signaling insufficient context. AdapShot then conducted a second round with 20, 24, 28, and 32 shots, achieving entropy values of 0.6862, 0.6884, 0.6232, and 0.6450. With both 28 and 32 shots falling below the threshold, AdapShot efficiently selected 28 shots as the optimal configuration. The model subsequently employed chain-of-thought reasoning to correctly extract \textit{1868} from the context. 

\begin{figure}[h]
\centering
\includegraphics[width=\linewidth]{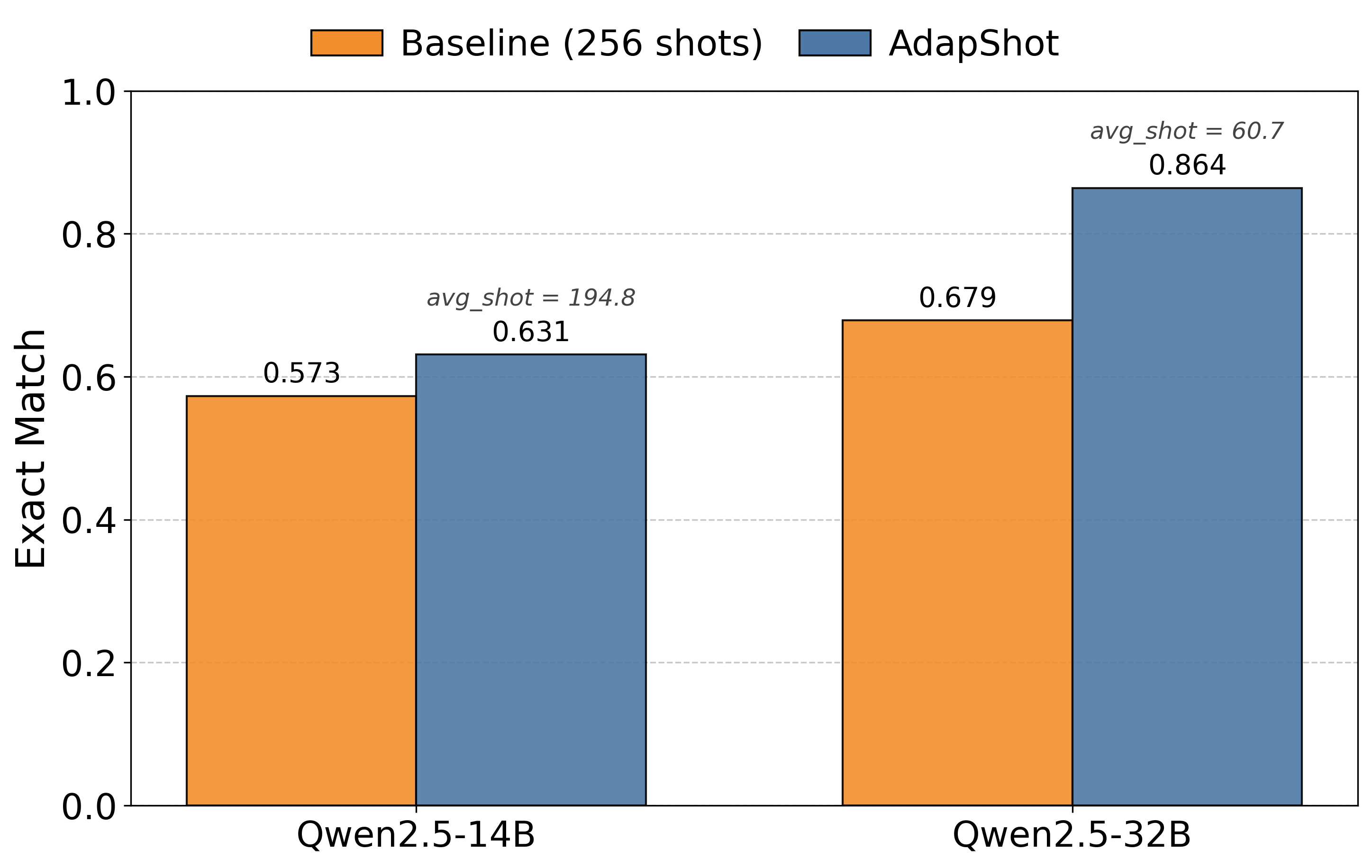}
\caption{Scalability analysis of AdapShot.
}
\label{fig:scaling_results}
\end{figure}

\textbf{Scaling with LLM Parameters:} We evaluated scalability on Qwen2.5-14B and 32B using the CoLA dataset. As shown in Figure \ref{fig:scaling_results}, AdapShot's advantage amplifies with model scale. Notably, on the 32B model, AdapShot achieves 86.41\% Exact Match using only $\sim$60 shots, whereas the fixed 256-shot baseline degrades significantly to 67.91\%. This sharp divergence confirms that larger models are highly sensitive to noise from irrelevant contexts. AdapShot effectively mitigates the heightened sensitivity of larger models to context noise by activating internal knowledge with a minimal, optimal set of demonstrations. 

\begin{table}[h]
\centering
\caption{Compare with BM25 based Many-Shot ICL.}
\label{tab:accuracy_bm25_comparison}
\resizebox{\linewidth}{!}{
\begin{tabular}{l|ccc}
\toprule
\textbf{Method} & \textbf{GSM8K} & \textbf{PIQA} & \textbf{SQuAD v2} \\
\midrule
BM25 + 256-shot & 0.456 & 0.379 & 0.418 \\
BM25 + 512-shot & 0.301 & 0.388 & 0.126    \\
\midrule
\textbf{AdapShot (Ours)} & \textbf{0.649} & \textbf{0.599} & \textbf{0.465} \\
\bottomrule
\end{tabular}}
\end{table}
\textbf{Comparison with BM25-Baseline:} Table \ref{tab:accuracy_bm25_comparison} validates the robustness of AdapShot by comparing it against baselines utilizing BM25 for example retrieval. 
AdapShot outperforms the best BM25 configuration by significant margins (+0.193 on GSM8K, +0.211 on PIQA, and +0.047 on SQuAD v2). These results underscore that fixed-shot retrieval strategies are suboptimal compared to AdapShot, proving that the adaptive determination of shot count is a crucial factor.

Our AdapShot is not sensitive to specific probe threshold values ($\tau$) and maintains consistent effectiveness.

\begin{table}[h!]
\centering
\caption{Performance of \textbf{openPangu-Embedded-1B-V1.1} under different in-context learning shot settings.}
\label{tab:shot_performance_openpangu_s}
\resizebox{\linewidth}{!}{
\begin{tabular}{lcccc}
\toprule
\textbf{Method (Shots)} & \textbf{ARC-Easy} & \textbf{CoLA} & \textbf{PIQA} & \textbf{QNLI}  \\
\midrule
Baseline (0-shot)   & 0.687 & 0.502 & 0.221 & 0.679 \\
Baseline (8-shot)   & 0.679 & 0.512 & 0.246 & 0.699\\
Baseline (128-shot) & 0.182 & 0.413 & 0.172 & 0.546 \\ 
\midrule
\textbf{AdapShot (Ours)} & 0.704 & 0.650 & 0.271 & 0.694 \\
\bottomrule
\end{tabular}
}
\end{table}

\textbf{Performance Comparison on OpenPangu:} To further evaluate the generalizability of the proposed AdapShot method, we conduct additional experiments on the \textbf{openPangu-Embedded-1B-V1.1} \cite{chen2025pangu}. Table~\ref{tab:shot_performance_openpangu_s} details the performance comparison including ARC-Easy, CoLA, PIQA, and QNLI. 
AdapShot outperforms the fixed-shot baselines on most tasks and achieves the better performance, without requiring manual shot selection.

\begin{table}[h!]
\centering
\caption{Inference speedup of AdapShot on \textbf{openPangu-Embedded-1B-V1.1}.}
\label{tab:main_efficiency_pangu_s}
\resizebox{\linewidth}{!}{
\begin{tabular}{l|ccccc}
\toprule
\textbf{Dataset} 
& \textbf{vs. 8-shot} 

& \textbf{vs. 64-shot} 
& \textbf{vs. 128-shot} \\
\midrule
ARC-Easy 
& 1.47$\times$ & 2.03$\times$ & 3.50$\times$  \\
GSM8K 
&  3.88$\times$  & 5.07$\times$ & 6.73$\times$   \\
PIQA 
& 1.71$\times$ & 1.56$\times$ & 2.13$\times$    \\
QNLI 
&  1.23$\times$  & 1.57$\times$ & 2.78$\times$  \\
\bottomrule
\end{tabular}
}
\end{table}
Table~\ref{tab:main_efficiency_pangu_s} illustrates the computational efficiency of AdapShot compared with the many-shot baseline on the openPangu architecture. 
Overall, our method delivers speedup across all evaluated scenarios, confirming that AdapShot provides both superior predictive performance and significant latency reductions.

%% file: sections/conc.tex
\section{Conclusion}
This paper proposes AdapShot, 
which utilizes a probing-based mechanism to assess query difficulty and combines semantic-aware KV cache reuse with position-decoupled re-encoding techniques to dynamically match the optimal number of shots for each query. This method effectively eliminates repetitive context prefilling computations, significantly reducing redundant overhead while improving operational efficiency. Experiments demonstrate that AdapShot reduces latency while boosting performance. In the future, we will explore implicit shot generation techniques to reduce reliance on real data.

\section{Limitations}
Although AdapShot demonstrates superior performance in improving inference efficiency and dynamically adapting the number of shots, this study still presents certain limitations. Specifically, similar to traditional many-shot ICL methods, AdapShot relies on retrieving or constructing real data samples to serve as context demonstrations. This process inevitably incurs additional data storage overhead and requires significant effort in sample curation. Therefore, we believe a promising direction for future work is to explore the synthesis of "implicit shots" to replace the current paradigm of explicitly constructing prompts with real samples, thereby further reducing reliance on real-world data and enhancing generalizability.

\section{Acknowledgments}
This work is supported by the National Key R\&D Program of China (No. 2026YFE0199800), the Chengdu Science and Technology Bureau Project (No. 2024-YF09-00041-SN), the National Natural Science Foundation of China Project with ID W2433163, the Sichuan Science and Technology Program (Grant No. 2026NSFSC1474), the Postdoctoral Fellowship Program (Grade C) of the China Postdoctoral Science Foundation (Grant No. GZC20251053) and the UESTC Kunpeng \& Ascend Center of Cultivation (Project ID: H04W241592).